\documentclass[a4paper]{cas-dc}
\usepackage[linesnumbered,ruled,vlined]{algorithm2e} 
\usepackage{subcaption}
\usepackage{multirow}
\usepackage{graphicx}
\usepackage{xcolor}
\usepackage{subcaption}
\usepackage{balance}
\usepackage{epsfig,endnotes}
\usepackage{import}
\usepackage{url}            
\usepackage{commath}
\usepackage{booktabs}       
\usepackage{amsfonts}       
\usepackage{nicefrac}       
\usepackage{microtype}      
\usepackage{array}
\usepackage{stfloats} 
\usepackage{epigraph}
\usepackage[switch,columnwise]{lineno}
\usepackage{setspace}

\usepackage[authoryear]{natbib}
\let\printorcid\relax

\SetKwInput{KwInput}{Input}               
\SetKwInput{KwOutput}{Output}

\def\name{OliVaR\xspace}

\begin{document}

\let\WriteBookmarks\relax


\shorttitle{\name}
\shortauthors{Miho et~al.}
\title[mode = title]{\name: Improving Olive Variety Recognition using Deep Neural Networks}                       
\author[1]{Hristofor Miho}
\ead{hmiho@uco.es}
\credit{Conceptualization of this study, Methodology, Writing - Original draft preparation, Dataset Creation}

\author[2]{Giulio Pagnotta}
\ead{pagnotta@di.uniroma1.it}
\credit{Conceptualization of this study, Methodology, Writing - Original draft preparation, Software}

\author[2]{Dorjan Hitaj}
\ead{hitaj.d@di.uniroma1.it}
\credit{Conceptualization of this study, Methodology, Writing - Original draft preparation}

\author[2]{Fabio {De Gaspari}}
\ead{degaspari@di.uniroma1.it}
\credit{Conceptualization of this study, Methodology, Writing - Original draft preparation}

\author[2]{Luigi Vincenzo Mancini}
\ead{mancini@di.uniroma1.it}
\credit{Conceptualization of this study, Methodology, Writing - Original draft preparation}

\author[3]{Georgios Koubouris}
\ead{koubouris@elgo.gr}
\credit{Methodology, Writing - Original draft preparation, Dataset Creation}

\author[4]{Gianluca Godino}
\ead{gianluca.godino@crea.gov.it}
\credit{Methodology, Dataset Creation}

\author[5]{Mehmet Hakan}
\ead{mehmet.hakan@tarimorman.gov.tr}
\credit{Methodology, Dataset Creation}

\author[1]{Concepcion Muñoz Diez}
\ead{cmdiez@uco.es}
\credit{Methodology, Writing - Original draft preparation, Dataset Creation}

\affiliation[1]{organization={Universidad de Cordoba},
    city={Cordoba},
    country={Spain}}
    
\affiliation[2]{organization={Sapienza Università di Roma, Dipartimento di Informatica},
    city={Rome},
    country={Italy}}

\affiliation[3]{organization={H.A.O. DEMETER, NAGREF, Institute of Olive Tree, Subtropical Crops and Viticulture},
    city={Chania, Crete},
    country={Greece}}

\affiliation[4]{organization={Council for Agricultural Research},
    city={Rende},
    country={Italy}}
    
\affiliation[5]{organization={Olive Research Institute},
    city={Izmir},
    country={Turkey}}


\maketitle
\begin{abstract}
The easy and accurate identification of varieties is fundamental in agriculture, especially in the olive sector, where more than 1200 olive varieties are currently known worldwide. Varietal misidentification leads to many potential problems for all the actors in the sector: farmers and nursery workers may establish the wrong variety, leading to its maladaptation in the field; olive oil and table olive producers may label and sell a non-authentic product; consumers may be misled; and breeders may commit errors during targeted crossings between different varieties. To date, the standard for varietal identification and certification consists of two methods: morphological classification and genetic analysis. The morphological classification consists of the visual pairwise comparison of different organs of the olive tree, where the most important organ is considered to be the endocarp. In contrast, different methods for genetic classification exist (RAPDs, SSR, and SNP). Both classification methods present advantages and disadvantages. Visual morphological classification requires highly specialized personnel and is prone to human error. Genetic identification methods are more accurate but incur a high cost and are difficult to implement.

This paper introduces \name, a novel approach to olive varietal identification. \name uses a teacher-student deep learning architecture to learn the defining characteristics of the endocarp of each specific olive variety and perform classification.
We construct what is, to the best of our knowledge, the largest olive variety dataset to date, comprising image data for 131 varieties from the Mediterranean basin. We thoroughly test \name on this dataset and show that it correctly predicts olive varieties with over 86\% accuracy.

\end{abstract}
\section{Introduction}\label{sec:introduction}

The olive tree (\textit{Olea europaea} L.) represents a priceless genetic variability heritage with more than 1200 varieties worldwide selected over more than 5500 years of cultivation. Due to its unique characteristics, this crop is an inherent part of the Mediterranean culture and mythology~\cite{Rallo2018a,Rugini2016}. On the other hand, the olive's high genetic variability contributed to a wide range of derived products~\cite{Miho2021,Rallo2018}.
Nowadays, olive genetic resources are conserved by a network of 23 national and international Germplasm Banks (GBs) coordinated by the International Olive Council - (“International Olive Council - Germplasm Banks Network,” 2020). The olive oil and table olive trade has experienced a recent market boom~\cite{GlobalTrade2021}. Consumers are increasingly interested in healthy food to improve their quality of life and prevent chronic diseases~\cite{LopezMirandaLancet, Casini2014,LuisaBadenes2012}.
The identification of olive varieties is a complex and crucial process that affects all stakeholders and end-users. Hence, considerable scientific efforts are invested in the development of new methods able to perform an efficient and reliable identification ~\cite{Atienza2013,Barranco2000,Belaj2022,Trujillo2014,Rugini2016}. The accurate identification of varieties guarantees the correct management of germplasm banks, the distribution and marketing of true-to-type varieties by nurseries, fair trading and the consumer confidence~\cite{Bartolini2005,Haouane2011,Koubouris2019}. The table and olive oil label “Protected Designation of Origin” (PDO) is among the most demanded by consumers, as it is associated with organoleptic quality and nutritional properties~\cite{Parra-Lopez2015}. Therefore, the proper identification of olive varieties is essential for numerous reasons but it is a complex and time-consuming task, requiring specialized personnel and expensive equipment~\cite{Likudis2016, SatorresMartinez2018}. 
The most widespread and community-accepted methods for olive varietal identification are based on the application of morphological and genetic markers~\cite{Trujillo2014}. 

Morphological markers in olives were firstly selected and applied for varietal classification in 1984 by Barranco et al.~\cite{Barranco1984}. In the 2000s, a simplified morphological scheme proposed by Barranco et al.~\cite{Barranco2000} was adopted as the reference by the International Union for the Protection of New Varieties of Plants (UPOV), which is still in force today~\cite{UPOV2011}. This pomological scheme allowed the cataloging of 272 Spanish olive varieties~\cite{Barranco2005}. It includes 24 characters describing the tree (3), leaf (4), fruit (7), and endocarp (10). Out of these 24 characters, those of the endocarp (olive pit) were considered the most meaningful for the varietal identification. Indeed, the endocarp could be considered the natural fingerprint of the olive tree~\cite{Hannachi2017}. The endocarp is the principal organ for varietal identification because: a) environmental factors scarcely influence its morphology; b) it presents significant polymorphism among varieties; c) it is to preserve, and transport ~\cite{Barranco2005}; and d) and its analysis presents a low implementation cost~\cite{Laaribi2017}. However, despite all these benefits, performing an accurate and reliable morphological characterization of olive varieties requires thorough training, being highly prone to human error ~\cite{SatorresMartinez2018,Sun2016}.

On the other hand, molecular techniques for olive cultivar identification were developed in the 1990s~\cite{Belaj2003, Trujillo1995}. The first genetic markers applied for varietal identification were Random Amplified Polymorphic DNA (RAPDs)~\cite{Belaj2003}. Later on, these markers were replaced by Microsatellites or Simple Sequence Repeat markers (SSRs), which demonstrated a robust discrimination capacity thanks to their large polymorphism. SSRs in combination with morphological markers, have been widely implemented, giving robust and useful results in the identification of olive cultivar collections~\cite{Emmanouilidou2018,Trujillo2014}. Genetic markers provide higher discriminatory capacity than the morphological markers. However, some limitations have been observed related to SSR markers, such as the complexity of establishing clear thresholds for intra- and inter-varietal variability~\cite{Bakkali2019,Baldoni2009,Trujillo2014}. Also, in a few cases, phenotypically different accessions presented the same or very similar SSR profile, leading to their classification as different varieties ~\cite{Barranco2005}. In addition, the International Union for the Protection of New Varieties of Plants (UPOV) primarily credits the morphological rather than genetic characterization. Therefore, morphological characterization is mandatory for the technical examinations of distinctness, uniformity, and stability required to register a new variety ~\cite{UPOV2011}.
Recently, Single-Nucleotide Polymorphism (SNP) markers joined the list of genetic markers for the varietal identification of olive trees. These markers are described as more powerful than the above-mentioned genetic markers, reducing the error rate of genotyping~\cite{Belaj2018,Belaj2022}. However, the bottleneck of using genetic tools on a large scale is still their high cost, time consumption, and the need for qualified human resources and sophisticated equipment.

This paper introduces \name, a deep learning olive variety recognizer based on endocarp photos. \name uses knowledge-driven learning paradigm to learn the defining characteristics of the endocarp of each specific olive variety and perform classification.
We construct a large-scale dataset of olive endocarp photos, comprising over 72,000 pictures from 131 different olive varieties, and show that \name can reliably classify them with over 86\% accuracy. 

To summarize, the contributions of this paper are the following:
\begin{itemize}
    \item We introduce \name, a deep learning model based on the morphological characteristics of the olive endocarp for varietal classification, thereby automatizing the traditional process of morphological classification.
    \item We construct a large-scale dataset of over 72,000 olive endocarp photos spanning 131 varieties from 4 of the largest olive germplasm banks of the Mediterranean area. To our knowledge, this is the largest dataset of olive endocarp photos to date.
    \item We thoroughly evaluate \name on this dataset and show that it is able to recognize olive varieties with high accuracy.
    \item We perform an analysis of what features of the endocarp \name focuses on, as well as a comparison between our proposed architecture and a state-of-the-art image recognition neural network.
\end{itemize}

This paper is organized as follows: Section~\ref{sec:background} provides relevant background knowledge necessary to understand the contributions. Section~\ref{sec:olivar} introduces \name, our DL-based olive variety recognizer. Furthermore, in Section~\ref{sec:exp_setup} we provide details about the experimental setup. Section~\ref{sec:evaluation} provides the \name evaluation results and discussion on the findings. In Section~\ref{sec:related_work} we discuss related work in the domain and Section~\ref{sec:conclusion} concludes the paper.
\section{Background}\label{sec:background}

\subsection{Deep Learning}\label{sec:background_dl}
Supervised machine learning algorithms make use of labeled data to produce a classifier that is able to predict the label of new, previously unseen instances. Given a set of independent variables $\mathbf{x}$, the machine learning model should predict a target outcome variable $y$. To do so, a function that maps these inputs $\mathbf{x}$ to the desired output $y$ needs to be learned.
This learning can be expressed using the following optimization problem:
    \begin{equation}
    \widehat{\theta}=\arg\min_{\theta\in \Theta} \sum_i l(f(\mathbf{x}_i; \theta) ,y_i),
    \label{eq:general_fr}
    \end{equation}
where $\hat{y}=f(\mathbf{x};\widehat{\theta})$ represents the learning machine. The learned function $f$ provides an estimate of the label $y$ for an input $\mathbf{x}$. The learning is guided by the loss function $l(\hat{y}, y)$ that measures the error for misclassifying $y$'s, providing useful information on how the parameters should be tuned in order for the learned machine to perform better on the task at hand. Typically machine learning algorithms are susceptible to overfitting. Overfitting occurs when the algorithm learns the training data ``too'' well (i.e., memorizing them), but this performance does not generalize well on unseen data. To cope with this issue the learning framework depicted on Equation~\ref{eq:general_fr} can be modified by adding an extra term $\Omega(\theta)$ which is independent of the training data~\ref{eq:general_fr2}.

    \begin{equation}
    \widehat{\theta}=\arg\min_{\theta\in \Theta} \sum_i l(f(\mathbf{x}_i; \theta) ,y_i) +\Omega(\theta),
    \label{eq:general_fr2}
    \end{equation}

Supervised learning algorithms such as Support Vector Machines (SVMs) \cite{Schoelkopf02}, Random Forests~\cite{Breiman01}, and deep neural networks (DNNs)~\cite{goodfellow2016deep} can be expressed using~\ref{eq:general_fr2}.
Deep Learning (DL) relies heavily on the use of Neural Networks (NN), which are machine learning (ML) algorithms inspired by the human brain and are designed to resemble the interactions amongst neurons~\cite{machinelearningMitchell}. While standard ML algorithms require the presence of handcrafted features to operate, NNs determine relevant features \emph{on their own}, learning them directly from the input data during the training process~\cite{goodfellow2016deep}.
Two main requirements underline the success of NNs in general: 1) large quantities of training data, and 2) powerful computational resources. Large amounts of diverse training data enable NNs to learn features suitable for the task at hand, while simultaneously preventing them from memorizing (i.e., overfitting) the training data. Such features are better learned when NNs have multiple layers, thus the \textbf{deep} neural networks. Research has shown that the single-layer, shallow counterparts are not good for learning meaningful features and are often outperformed by other ML algorithms~\cite{goodfellow2016deep}. DNN training translates to vast numbers of computations requiring powerful resources, with graphical processing units (GPUs) a prime example.
DL is the key factor for an increased interest in research and development in the area of Artificial Intelligence (AI), resulting in a surge of ML based applications that are reshaping entire fields and seedling new ones. Variations of DNNs, the algorithms residing at the core of DL, have successfully been implemented in multiple domains, including here, but not limited to, image classification~\cite{Simonyan14verydeep, He2016DeepRL, Chollet2017XceptionDL}, natural language processing~\cite{nlp1,nlp2,recommender_2}, speech recognition~\cite{speech2, speech3}, data (image, text, audio) generation~\cite{menick2018generating, styleGAN, iresnet, 9833616}, cyber-security~\cite{naked_sun,maleficnet,minerva}, and even aiding with the COVID-19 pandemic~\cite{lozano2021open}. 
\section{\name}\label{sec:olivar}

In this section, we present \name, our novel neural network-based olive variety recognition approach.
\name is constructed around a knowledge-driven learning (KDL) paradigm. KDL paradigm consists of boosting the learning capabilities of a model over a dataset by constructing an ensemble of models (which act as experts). These experts are used via transfer learning to guide the learning process of another deep neural network model. The KDL paradigm is shown to be able to assist in the training of an ML model by allowing it to a) converge faster and b) achieve good performance under conditions of limited training data or task complexity.
Both characteristics are fairly welcome given the task at hand. Especially the second, as the collection of the olive fruit, the specific genetic checks to guarantee the varietal authenticity and correct labeling, and the processing of the olive stones/endocarps to be photographed is a laborious task that requires significant efforts and costs. Therefore, in this study, we had to limit the number of photographed endocarps to 150-200 since we are evaluating about 131 different olive varieties distributed in the international Germplasm Banks (GBs) of 4 different countries (Spain, Italy, Greece, and Turkey).

We base the foundations of \name on recent work on KDL approach from~\cite{avola_1,avola_2}. Similar to Avola et al.~\cite{avola_1,avola_2}, our olive variety recognizer is based on three main components. Those components are the data augmentation component, the ensemble of experts and the knowledge-driven component. The latter two are considered as \name model architecture in what follows.

\subsection{Data augmentation component}
\begin{figure*}[t]
    \centering        
	    \begin{subfigure}{.3\textwidth}
            \centering
            \includegraphics[width=\columnwidth]{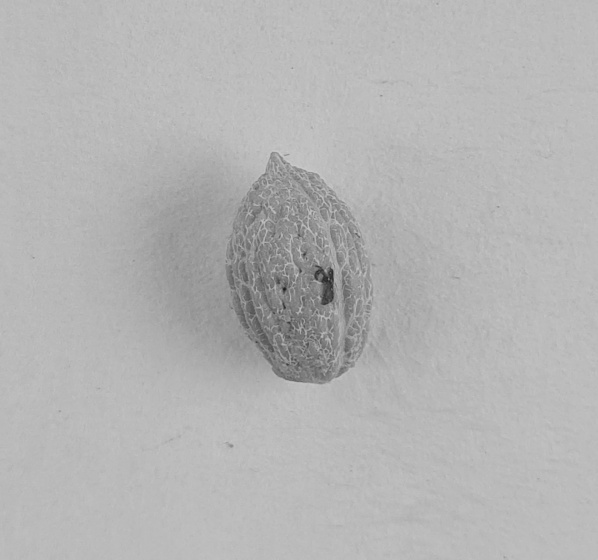}
             \caption{Grayscale}
             \label{fig:grayscale}
        \end{subfigure}
        \hfill
	    \begin{subfigure}{.3\textwidth}
            \centering
            \includegraphics[width=\columnwidth]{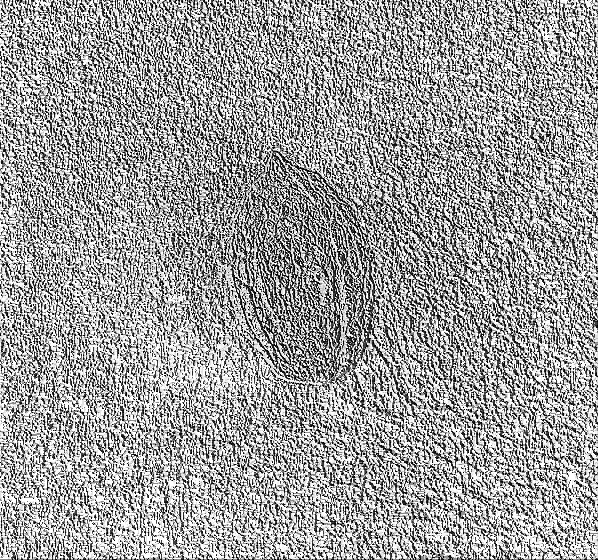}
             \caption{LBP}
             \label{fig:lbp}
        \end{subfigure}
                \hfill
	    \begin{subfigure}{.3\textwidth}
            \centering
            \includegraphics[width=\columnwidth]{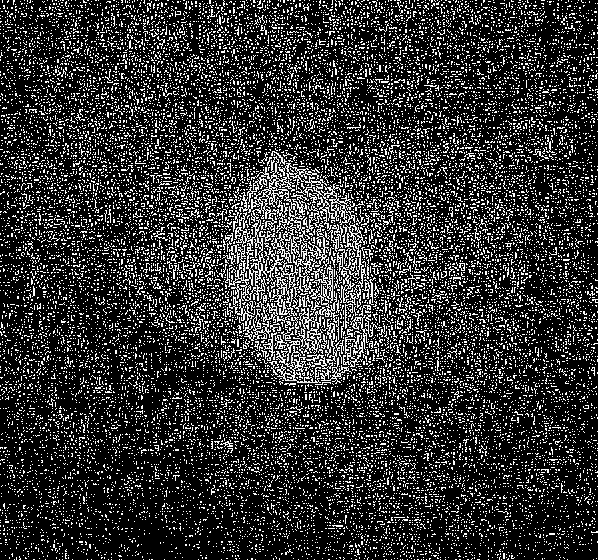}
             \caption{DWT}
             \label{fig:dwt}
        \end{subfigure}
    \caption{The dataset augmentation techniques employed in \name.} 
    \label{fig:datast_preprocessing}
\end{figure*}

\begin{figure*}[t]
  \centering
    \includegraphics[width=.6\textwidth]{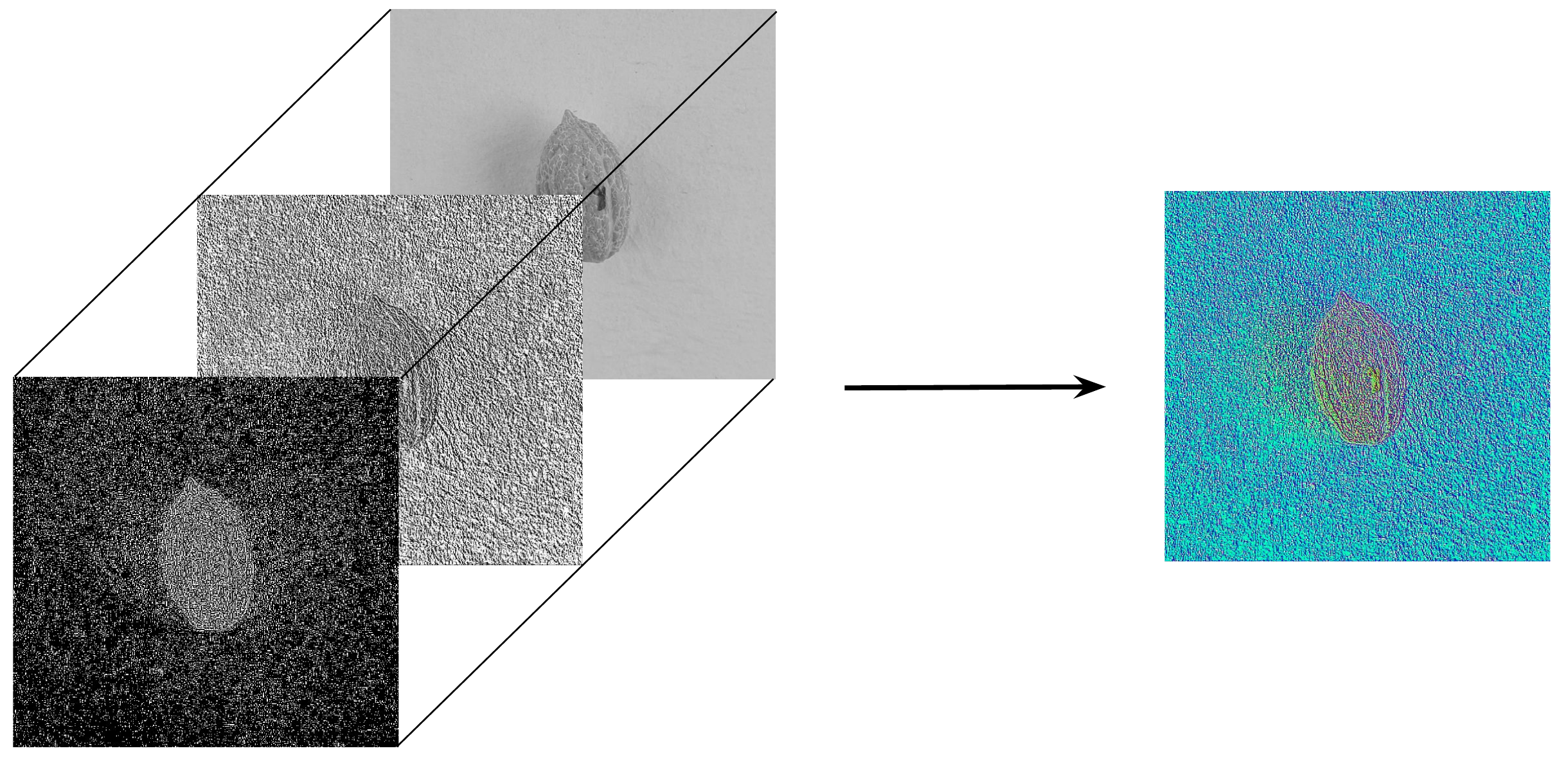}
    \caption{The resulting image after stacking all three individual images obtained in the dataset preprocessing phase.}\label{fig:image_stack}
 \end{figure*}
Given that we are attempting a KLD learning paradigm due to the scarcity of data and the complexity of the task, we first have to implement data augmentation techniques to extract more information from the data. To do so, we need to enhance the features of the olive endocarps, such as the texture of the surface of the olive, to help the model learn the small differences that occur between different olive varieties. In what follows, we present the preprocessing steps undertaken:

\begin{enumerate}
    \item First, we convert the image into a greyscale image to eliminate possible bias given by the colors that can be conditioned by the different lighting conditions that may have been present during the data collection. Once the grayscale image is obtained, we proceed with the data augmentation techniques we selected to enhance the peculiarities of the olive endocarp.  (as shown in Figure~\ref{fig:grayscale})
    
    \item Secondly, we use the Local Binary Pattern (LBP)~\cite{576366} augmentation technique on the grayscale version of the image. The LBP method is usually used to study the local properties of the image and identify the characteristics of individual parts of the image, such as textural information, using a combination of statistical and structural methods (as shown in Figure~\ref{fig:lbp}).
    
    \item Thirdly, we used Discrete Wavelet Transform (DWT) \cite{388960} on the grayscale version of the image. DWT has been successfully used in state-of-the-art applications for texture recognition, and it is instrumental in this particular task to highlight the texture of the olive endocarp surface (as shown in Figure~\ref{fig:dwt}).
\end{enumerate}

\begin{figure*}[t]
  \centering
    \includegraphics[width=.8\textwidth]{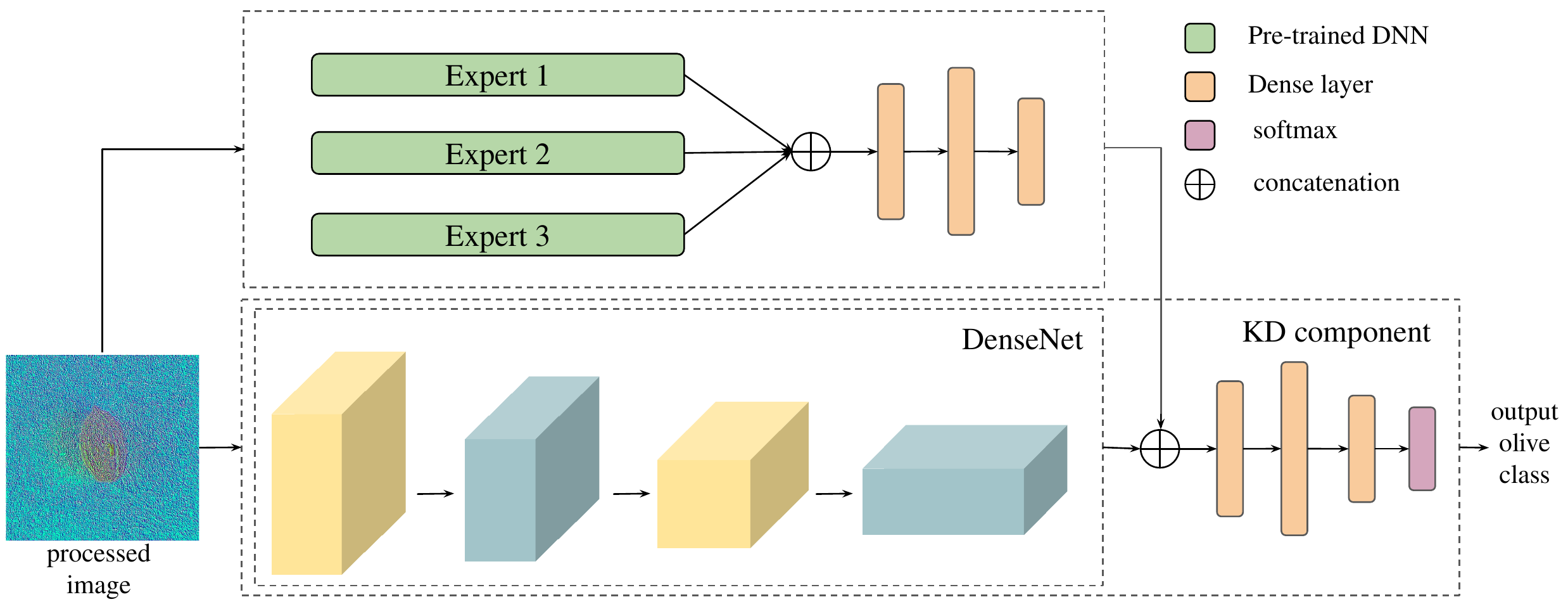}
    \caption{\name architecture composed of an ensemble of architectures of pre-trained models that are fine-tuned to learn the olive recognition task in order to provide assistance to a standalone neural network for better performance.}\label{fig:olivar_architecture}
 \end{figure*}
After obtaining the augmented images, we stack them together to obtain a three-channel image where, instead of the RGB channels, we have the grayscale image, the LBP-generated image, and the DWT-generated image corresponding to each individual channel.
We can do so because all images are grayscale, and thus this stacking procedure results in an image representing an olive endocarp with shape $(w \times h \times 3)$, where w and h correspond to width and height of the image, where each channel is a different representation of the olive endocarp (i.e., grayscale, DWT, LBP).
Once the images are stacked together, we obtain an image like the one depicted in Figure~\ref{fig:image_stack}. We process the whole dataset following this procedure.
We decided to choose the LBP and DWT representations (alongside the grayscale version of the olive endocarp) because they are shown to provide more information about the texture in small objects, thus resulting in better performance in such tasks~\cite{avola_1}.

\subsection{\name Architecture}

As previously mentioned, the \name architecture is composed of two main components, the ensemble of experts and the knowledge-driven component, which have been proven effective in other compelling computer vision tasks~\cite{avola_2}. Figure~\ref{fig:olivar_architecture} represents the high-level overview of \name architecture.
The ensemble of experts is composed of three pre-trained neural network architectures, defined \textit{experts} in prior works, that are fine-tuned to solve the olive variety recognition task. Each expert is modified, removing its original last layer and substituting it with a new dense layer. During the training, all the weights of the expert, but the new final dense layer, are frozen, i.e., only the last dense layer is updated. During the training, the experts are fine-tuned to recognize the 131 different olive varieties, and their predictions are concatenated and re-elaborated via a neural network composed of three dense layers. We highlight that while the experts' weights are frozen, except for the last layer, the weights of this neural network are constantly updated during the training to learn the best interpretation of the experts' predictions. This interpretation is combined with the prediction of the other component of our general architecture, which is a DenseNet. This DenseNet architecture is trained from scratch to recognize the different varieties based on the augmented olive endocarps images and the experts' prediction. Indeed, to compose the KD component of our architecture, the prediction of the DenseNet is concatenated with the experts' prediction and then passed through another neural network consisting of three dense layers.   
In section~\ref{sec:evaluation} we show that this custom architecture allows \name to attain a plausible performance in the olive variety recognition task, being able to classify over 131 different olive varieties.
\section{Experimental Setup}\label{sec:exp_setup}

\subsection{Dataset}\label{sec:dataset_details}
The dataset used to train \name consists of 72,690 images spanning 131 classes corresponding to 131 different olive varieties. The dataset was created in collaboration with four international olive germplasm banks that collected the fruits, ensuring the accuracy of the labeling of each fruit using SSR genetic markers.
For each variety, about 150-200 olive fruits were collected at maturity index over two once pit hardening and fruit formation has been consolidated~\cite{Rallo2018}. The endocarp was removed from the fruit flesh and carefully cleaned to ensure that the endocarp patterns were as clean and preserved as possible.

The endocarp cleaning procedure consisted of two main steps:
\begin{itemize}
    \item \textbf{Endocarp extraction}: The endocarp of the olives was extracted using a manual pitting machine. In order to smooth the endocarp extraction and cleaning process, the fruits underwent a freeze-thawing process to soften the flesh of the fruit.
    
    \item \textbf{Cleaning, bleaching and drying}: Once the endocarps were extracted from the fruits, the remaining fruit flesh was removed. To do this, we used a plastic mesh, and by scrubbing and rinsing with water, the flesh remains were cleaned off. This task was important because no flesh residue can remain in the endocarp, as it hinders its morphological characteristics, as well as to avoid the growth of fungus. The next step is bleaching. For this purpose, a 50\% solution of bleaching agent (e.g., sodium hypochlorite) is used. The endocarps are kept in sodium hypochlorite for 30 to 60 minutes until a clear whitish color is visible. Subsequently, the endocarps were dried at room temperature for one week or at 37ºC in the oven for 48 hours. Once dried, the endocarps were stored in labeled plastic containers indicating the name of the olive variety they belong.
\end{itemize}

\begin{figure}[t]
  \centering
    \includegraphics[width=.9\linewidth]{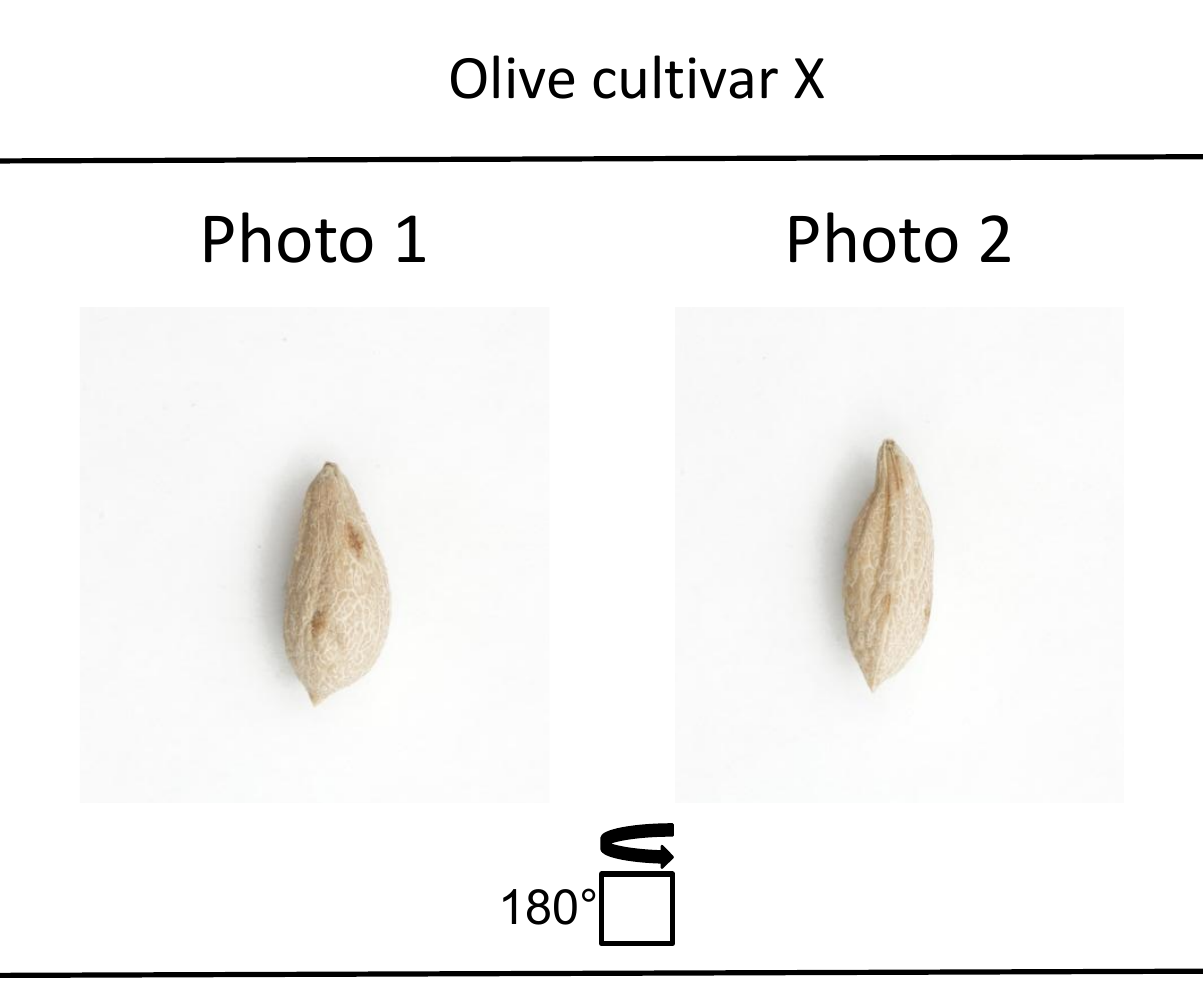}
    \caption{Dataset creation procedure. Each cleaned olive endocarp is photographed two times rotating it 180 degrees around its vertical axis.}\label{fig:dataset_creation}
 \end{figure}

After the cleaning and labeling procedures, two pictures were taken per endocarp. The first picture position has always been taken randomly, and the second position has been taken by rotating the first position by 180 degrees around its vertical axis (Figure~\ref{fig:dataset_creation}) to consider possible endocarp asymmetry. In this way, we build a dataset that will guide the ML model to generalize and learn to recognize the olive variety given an image corresponding to any side of the olive endocarp.

\subsection{Software and Hardware Requirements}\label{sec:hardware_requirements}
\name is built on top of version 1.7.1 of the PyTorch ML framework~\cite{pytorch}, using an environment with Python version 3.8.5. The experiments were conducted on a desktop PC running the Ubuntu 20.04.2 LTS operating system with a Ryzen 9 3900x processor, 64GB of RAM, and Nvidia GeForce RTX 2080Ti GPU with 11GB of memory.
\section{Evaluation}\label{sec:evaluation}
This section thoroughly evaluates the performance of \name on the test set and provides insights on the functionality of the approach.  

\subsection{Model Performance}
We trained \name using the training set described in section~\ref{sec:dataset_details}. Our results show that the performance of the custom architecture is superior to the performance of state-of-the-art architectures for image classification.

\begin{figure}[tb] 
\centering
 \includegraphics[width=.9\linewidth]{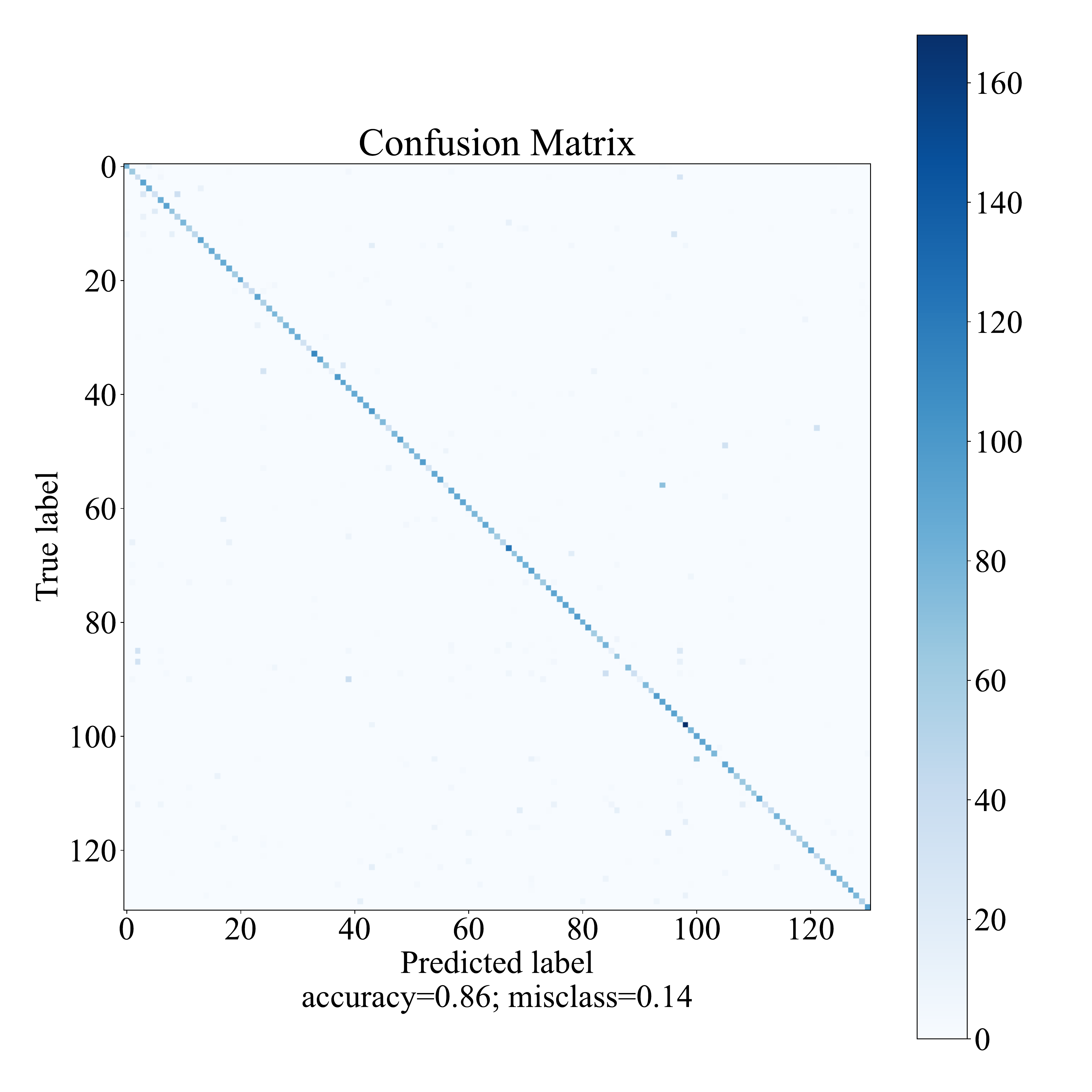}
\caption{\name confusion matrix over the test set. }
\label{fig:confusion_matrix}
\end{figure}

Figure~\ref{fig:confusion_matrix} represents the confusion matrix obtained by evaluation \name on the testing. The accuracy of \name was 86\% over the 131 olive varieties present in the test set. The y-axis of a confusion matrix represents the ground truth label for each test set sample, and the x-axis represents the label predicted by the model (\name in this case).
In a perfect scenario, only the square diagonal would be highlighted, meaning that the model could correctly predict each sample in the test set.
If we look at the diagonal in the confusion matrix, we see that only a few olive varieties are confused with each other, while most are correctly classified. These results demonstrate that a deep learning-based architecture is a good fit to distinguish olive varieties based on a photo of the endocarp, a method significantly less expensive than a DNA test.

\subsection{What has \name learned?}
\begin{figure}[tb] 
\centering
 \includegraphics[width=.9\linewidth]{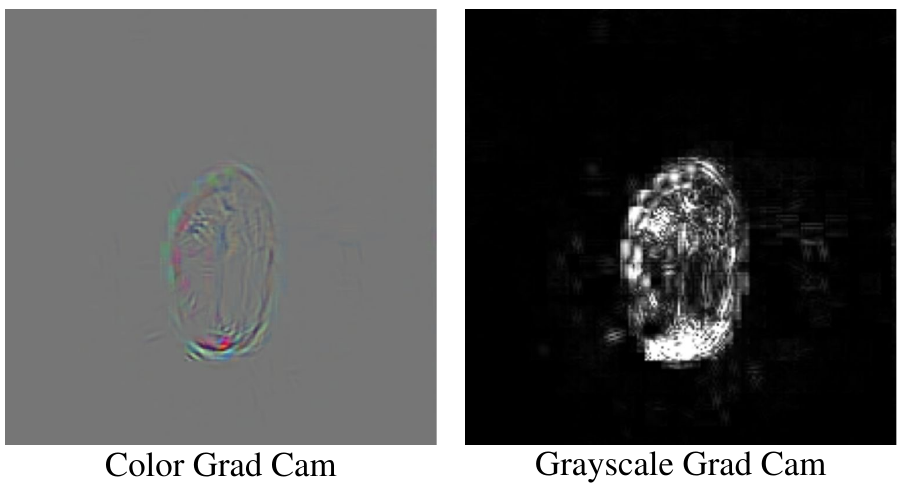}
\caption{Color Grad Cam and Grayscale Grad Cam output for sample of \textit{Bosana} olive variety.}
\label{fig:bosana_gc}
\end{figure}

\begin{figure}[tb] 
\centering
 \includegraphics[width=.9\linewidth]{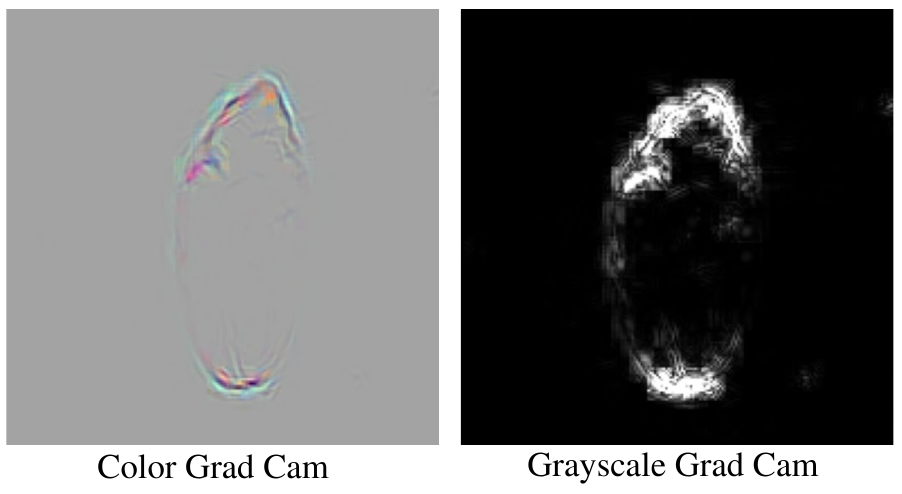}
\caption{Color Grad Cam and Grayscale Grad Cam output for sample of \textit{Mignolo Cerretano} olive variety.}
\label{fig:mignolo_gc}
\end{figure}

In this section, we delve deeper into the inner working of \name and try to evaluate how the model has learned and what regions of the image it considers as most significant in making a decision. To do so, we employ Gradient-weighted Class Activation Mapping (Grad CAM)~\cite{gradcamm}. 
Grad-CAM uses the gradients of any target concept in a classification network flowing into the final convolutional layer to produce a coarse localization map highlighting the important regions in the image that the network predominantly uses for predicting. We employ the Grad CAM technique in \name and in Figures~\ref{fig:bosana_gc} and~\ref{fig:mignolo_gc}, we present the output on samples from the test dataset, specifically for the olive varieties of \textit{Bosana} and \textit{Mignolo Cerretano}.
We notice in Figure~\ref{fig:bosana_gc} how \name focuses specifically on the olive endocarp region of the image, where we see the highlights of the olive endocarp pattern that \name uses to recognize that this particular olive endocarp belongs to the \textit{Bosana} olive variety. Interestingly, in this case the Grad-Cam is highlighting patterns all over the olive endocarp, which correspond to the rough pattern that \textit{Bosana} variety contains throughout the whole surface of the endocarp. On the contrary, in Figure~\ref{fig:mignolo_gc}, we notice how the Grad Cam output for a different olive variety, namely \textit{Mignolo Cerretano}, is different. We notice that Grad Cam has highlighted the upper and lower extremities of the olive endocarp. This means that \name is focusing more on those regions for this particular variety for classification. This is interesting to note due to the fact that the middle section surface of this particular olive variety is generally smooth and as such \name, during training has learned to focus more on the extremities. It is also interesting to note that morphological classification done by human experts also mostly focuses on the specific characters of the endocarp extremities~\cite{Barranco2000, Barranco2005, UPOV2011}.

\subsection{Discussion of findings}\label{sec:discussion}

Nowadays, when someone needs to solve an image classification task they commonly take a pretrained DNN architecture on the Imagenet~\cite{imagenet_cvpr09} task, typically being the ResNet~\cite{He2016DeepRL} or VGG16 and VGG19~\cite{Simonyan14verydeep} architectures and fine tune them on the dataset corresponding to the task at hand. Typically this approach works reasonably fine for most of the tasks. The task at hand that we treat in this work is a bit more complex to solve following this simplistic approach due to the limited amount of training data spread over a large number of classes.

\begin{figure}[tb] 
\centering
 \includegraphics[width=.9\linewidth]{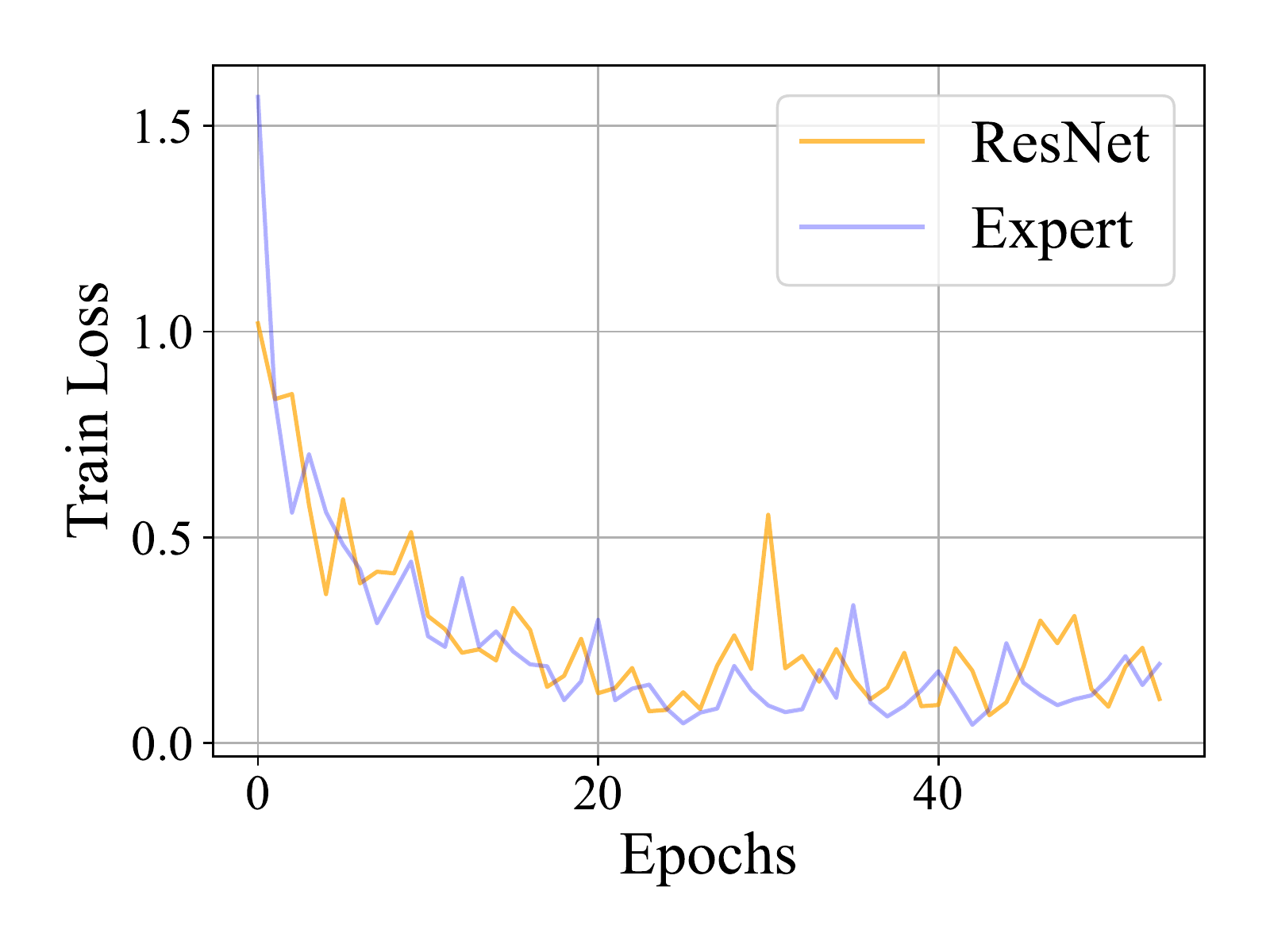}
\caption{Training loss comparison (ResNet vs. \name).}
\label{fig:train_loss_comparison}
\end{figure}

\begin{figure}[tb] 
\centering
 \includegraphics[width=.9\linewidth]{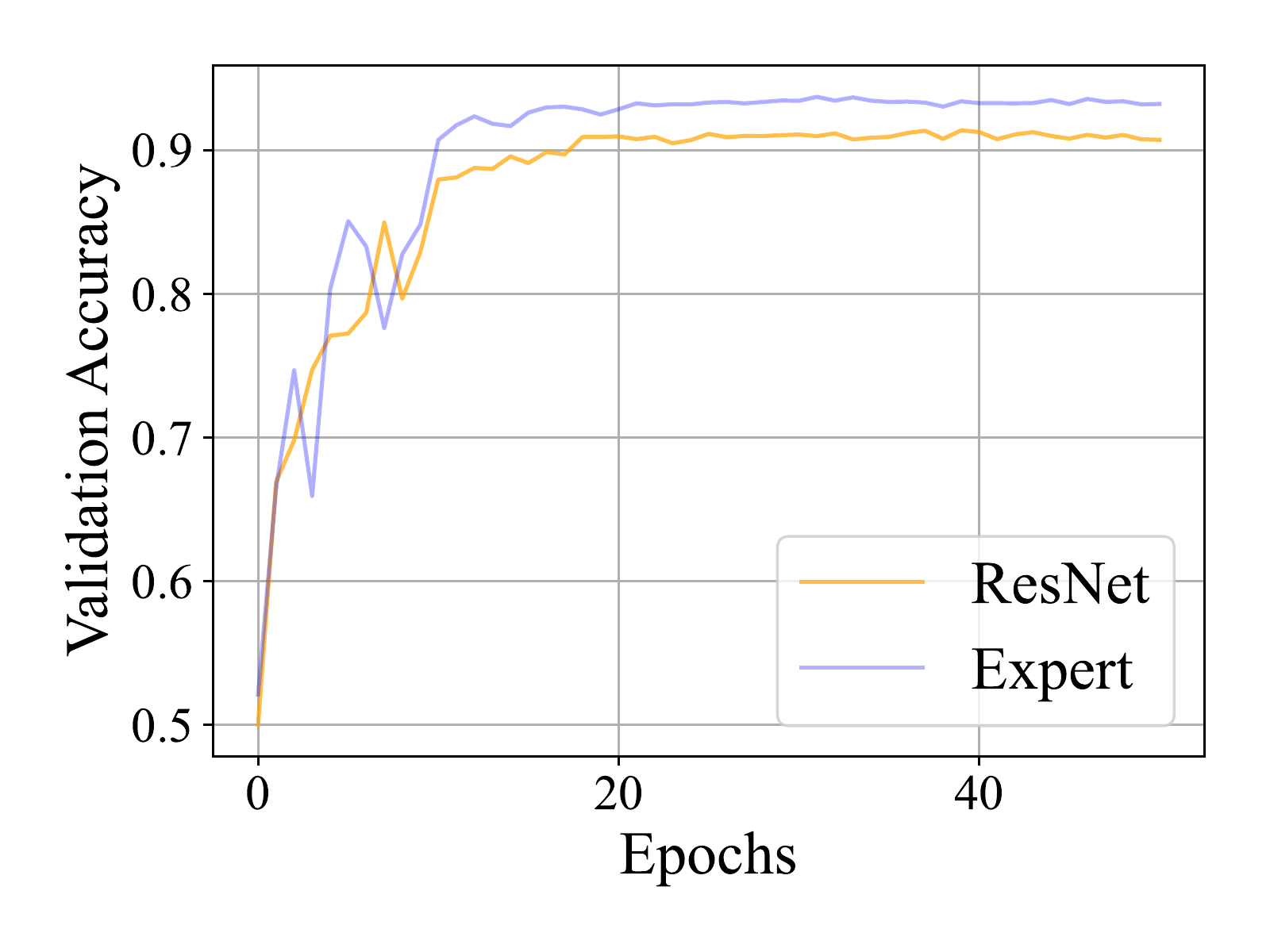}
\caption{Validation accuracy comparison (ResNet vs. \name).}
\label{fig:validation_accuracy}
\end{figure}

In this section we present a comparison between one state-of-the-art architecture, namely ResNet and our custom KDL architecture \name. In Figure~\ref{fig:train_loss_comparison} and Figure~ \ref{fig:validation_accuracy} we show how these two different architectures perform. In Figure~\ref{fig:train_loss_comparison} we note that the training loss of both ResNet and \name follows a similar pattern, needing the same amount of training epochs to converge. This is a plausible result due to the fact that \name is a much more complex architecture compared to ResNet, and requiring the same amount of time is satisfactory, especially when we look at the models performance. As shown in Figure~\ref{fig:validation_accuracy}, we see that the validation accuracy of \name is over three percentage points higher than ResNet highlighting the improvements that a KDL tailored architecture such as \name brings to this complex task.

\section{Related Work}\label{sec:related_work}

In the past decade, the scientific community has paid great attention and effort to computer technologies and statistical methods as an alternative or as a complement for carrying out varietal identification of the olive tree in a simpler, quicker, and more cost-effective way such as the works by~\cite{Avramidou2020,Koubouris2019,SatorresMartinez2018,Sesil2019,Vanloot2014}. The common goals of these investigations have been the automation and simplification of the traditional method of varietal morphological characterization based on the manual observation of the endocarp and other organs such as leaves.

To the best of our knowledge, prior to \name, \cite{Sesil2019} are the only ones that have attempted a DL-based approach for the olive varietal classification. Contrary to \name, the approach of Sesil et al. relies on olive leaves to perform the olive variety classification.
The approaches based on olive leaves are shown to be unreliable, and most prior work has considered the endocarp as the bearer of significant amounts of information for the varietal classification~\cite{Barranco2005,Koubouris2018,Koubouris2019}.
Furthermore, the approach by \cite{Sesil2019} is trained and evaluated on only four olive varieties compared to the 131 varieties on which \name is trained and evaluated, thus rendering \name a more complete tool for olive varietal classification.

Given that the olive endocarp is considered a more reliable source of information for varietal classification, a significant amount of work relying on computer and statistical techniques to automate or semi-automate the process of varietal identification through the endocarp morphological traits has been carried out. Specifically, \cite{Koubouris2019}, using the statistical method of Classification Binary Tree, correctly classified 42 olive varieties based on 11 endocarp traits previously extracted in a semi-automated way. 
In addition, statistical analysis of two-dimensional \cite{Koubouris2018} or three-dimensional images \cite{Manolikaki2022} have been successfully used to characterize 50 olive varieties. Similar and satisfactory results have also been obtained by other authors using statistical techniques such as Principal Component Analysis and Partial Least Square-discriminant analysis \cite{Blazakis2017, SatorresMartinez2018, Vanloot2014}.
However, these methods have not found widespread use by the olive growing community and the authorities due to the complexity of transferring the methods across entities \cite{Koubouris2019} and the high cost of intermediate steps in semi-automatic feature extraction techniques. Nevertheless, all these studies stress the fact that the morphological characteristics of the olive endocarp are undoubtedly a reliable fingerprint for varietal identification.

ML has been successfully introduced in oliviculture and has shown promising results. Specifically,~\cite{Khosravi2021} built image-based models to automatically estimate the fruit ripening stages, while~\cite{Cruz2007} have been able to predict or detect with high accuracy Xylella Fastidiosa. In a similar approach,~\cite{Diaz2004} developed models to classify the table olives into different quality categories, depending on the skin defects, with an accuracy of more than 90\%. 

Furthermore, DL has also been implemented on a massive scale in other plant species for a multitude of reasons, such as crop management, including applications in yield prediction, disease detection, weed detection, crop quality, and species recognition~\cite{Ali2017, Hussain2022, Liakos2018, Ramos2017, Sengupta2014}. Regarding varietal identification/classification in other species using DL models based on morphological characters, different authors have reported classification results showing a very high accuracy (between 90 and 98\%) as in the case of three plum varieties classification~\cite{Ropelewska2022}; the classification of 16 grapevine varieties~\cite{Fuentes2018}, classification of Durian varieties~\cite{Lim2019}; and classification of three legume varieties via leaf vein pattern analysis~\cite{Grinblat2016}.
\section{Conclusion}\label{sec:conclusion}
In this work, we presented \name, a neural network-based approach that is able to recognize olive varieties based on photos of the endocarp with an accuracy of 86\% over the 131 olive varieties considered.
\name outperforms human beings in this task, while having a close performance to the DNA-based olive variety recognition. Unlike DNA-based olive recognition, which typically requires days, expensive equipment, and specialized personnel to obtain the result, \name can provide an answer in just a few milliseconds.

We believe that \name will assist everyone involved in the olive sector, because the quick and accurate authentication of olive varieties is critical to avoid mistakes in establishing olive plantations or crossbreeding within a breeding program to obtain new varieties with better characteristics. An error in the varietal determination may lead to significant economic losses for a farmer or breeder.
Furthermore, the rapid detection of olive varieties can help the oil extraction industry to quickly authenticate and differentiate mono-varietal oils and avoid possible fraud for the end consumers.
However, this is the first preliminary work of such dimensions related to olive varietal identification. The results still need to be validated in time and space to prove the reproducibility of the model for further commercial use.

\section{Acknowledgments}\label{sec:acks}
This work was supported by GEN4OLIVE, a project that has received funding from the European Union’s Horizon 2020 research and innovation programme under grant agreement No. 101000427. The research has also been supported by the postdoctoral grant "Margarita Salas" (UCOR01MS, BOUCO n.º 2021/00729), awarded by the University of Córdoba (grants to Public Universities for the requalification of the Spanish university system from the Ministry of Universities, Spain) and funded by the European Union – NextGenerationEU.

Moreover, we are grateful for the generous contribution of the following entities and persons who, in the framework of GEN4OLIVE project, provided the olive endocarp photos for the development of \name:
\begin{itemize}
    \item University of Cordoba, Cordoba, Spain: Isabel Trujillo Navas, Diego Barranco Navero and Anna-Maria Volakaki.
    \item Institute of Olive Tree, Subtropical Crops and Viticulture, Crete, Greece: Ioanna Kaltsa.
    \item Olive Research Institute, Izmir, Turkey: Melek Gurbuz, Hulya Kaya.
    \item Council for Agricultural Research, Rende, Italy: Enzo Perri, Rosa Nicoletti and Annamaria Lenco.
    \item Institut National de la Recherche Agronomique, Marrakech, Morocco: Sara Oulbi.
\end{itemize}


\bibliographystyle{model1-num-names.bst}

\bibliography{bibliography}
\end{document}